%% file: main.tex
\documentclass[10pt,twocolumn,letterpaper]{article}

\usepackage{multirow}
\usepackage{soul}
\usepackage[section]{placeins}
\usepackage[pagenumbers]{cvpr} % To force page numbers, e.g. for an arXiv version


\definecolor{cvprblue}{rgb}{0.21,0.49,0.74}
\usepackage[pagebackref,breaklinks,colorlinks,allcolors=cvprblue]{hyperref}

\title{HIIF: Hierarchical Encoding based Implicit Image Function for Continuous Super-resolution}%{Vivid Vision Representation for Implicit Super-Resolution of Arbitrary Scale}

\author{
Yuxuan Jiang$^1$,
Ho Man Kwan$^1$,
Tianhao Peng$^1$,
Ge Gao$^1$,
Fan Zhang$^1$,  \\
Xiaoqing Zhu$^2$,
Joel Sole$^2$,
and David Bull$^1$ \\
$^1$ \textit{Visual Information Laboratory, University of Bristol, Bristol, BS1 5DD, UK}\\
$^1$ \textit{\{yuxuan.jiang, hm.kwan, tianhao.peng, ge1.gao, fan.zhang, dave.bull\}@bristol.ac.uk}\\
$^2$ \textit{Netflix Inc., Los Gatos, CA, USA, 95032}\\
$^2$ \textit{\{xzhu, jsole\}@netflix.com}\\
}

\begin{document}
\maketitle
\input{sec/0_abstract}    
\input{sec/1_intro}
\input{sec/2_related_work}
\input{sec/3_Method}

\input{sec/4_Experiment}
\input{sec/5_Conclusion}

\section*{Acknowledgements}
The authors appreciate the funding from Netflix Inc., the University of Bristol, and the UKRI MyWorld Strength in Places Programme (SIPF00006/1).

{
    \small
    \bibliographystyle{ieeenat_fullname}
    \bibliography{main}
}

% WARNING: do not forget to delete the supplementary pages from your submission 
% \input{sec/X_suppl}
% {
%     \small
%     \bibliographystyle{ieeenat_fullname}
%     \bibliography{sub}
% }

\end{document}

%% file: sec/0_abstract.tex
\begin{abstract}
Recent advances in implicit neural representations (INRs) have shown significant promise in modeling visual signals for various low-vision tasks including image super-resolution (ISR). INR-based ISR methods typically learn continuous representations, providing flexibility for generating high-resolution images at any desired scale from their low-resolution counterparts. However, existing INR-based ISR methods utilize multi-layer perceptrons for parameterization in the network; this does not take account of the hierarchical structure existing in local sampling points and hence constrains the representation capability. In this paper, we propose a new \textbf{H}ierarchical encoding based \textbf{I}mplicit \textbf{I}mage \textbf{F}unction for continuous image super-resolution, \textbf{HIIF}, which leverages a novel hierarchical positional encoding that enhances the local implicit representation, enabling it to capture fine details at multiple scales. Our approach also embeds a multi-head linear attention mechanism within the implicit attention network by taking additional non-local information into account. Our experiments show that, when integrated with different backbone encoders, HIIF outperforms the state-of-the-art continuous image super-resolution methods by up to 0.17dB in PSNR. The source code of HIIF will be made publicly available at \url{www.github.com}.
\end{abstract}

%% file: sec/1_intro.tex
\section{Introduction}
\label{sec:intro}

Image Super-Resolution (ISR) is an important research area in computer vision and graphics, focusing on the reconstruction of high-resolution images from their down-sampled low-resolution versions. The challenging nature of ISR, which arises from the need to recover fine details based on limited information, makes it an enduring problem in the field of low-level vision. Recent ISR research contributions leverage deep learning techniques, which have significantly advanced the state of the art in recent years \cite{dong2015image, kim2016accurate, lim2017enhanced, zhang2018image, ma2020mfrnet, liang2022vrt, wang2022uformer, liang2021swinir, conde2022swin2sr, jiang2024mtkd}. Most of these approaches perform up-sampling through sub-pixel convolution and train a static model for a single scaling factor; this constrains use cases, especially when the target resolution is unknown. To address this issue, arbitrary and continuous scale ISR methods have been proposed which enable a single model to achieve up-sampling with multiple and/or continuous scales \cite{wang2021learning, hu2019meta}. This enhances model generalization and flexibility and has attracted significant attention in recent years.

\begin{figure}
    \centering
    \begin{minipage}{1\linewidth}
		  \centering
            \setlength{\abovecaptionskip}{0.cm}
		  \includegraphics[width=1\linewidth]{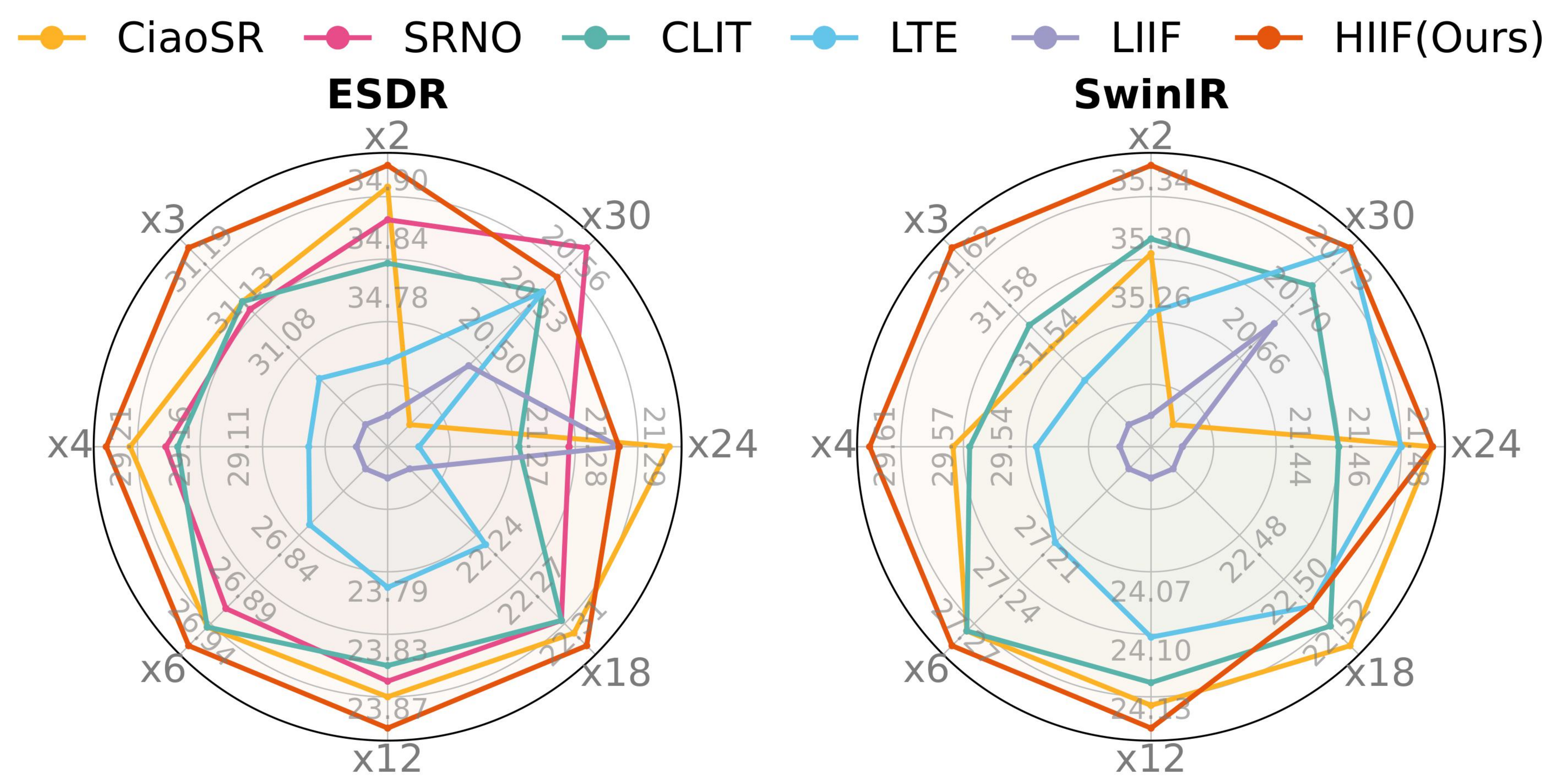}
          % \text{(a)}\vspace{.1cm}
		% \caption*{GT \protect\\ (PSNR/SSIM)}
	\end{minipage}

    \begin{minipage}{1\linewidth}
		  \centering
            \setlength{\abovecaptionskip}{0.cm}
		  \includegraphics[width=1\linewidth]{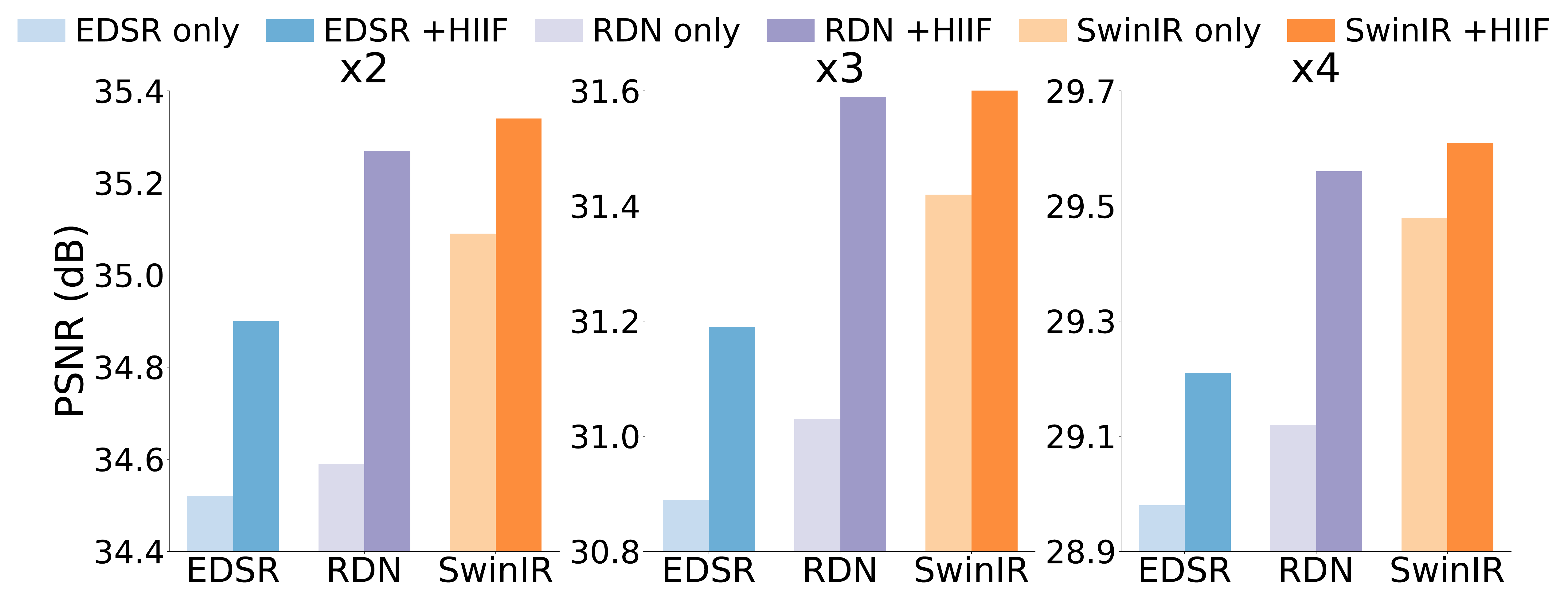}
          % \text{(b)}\vspace{.1cm}
		% \caption*{GT \protect\\ (PSNR/SSIM)}
	\end{minipage}
    \caption{(Top) Radar plots illustrating the performance of proposed HIIF and five other INR-based continuous ISR methods. (Bottom) HIIF versus ISR methods (based on the same architectures) with fixed in-distribution up-sampling scales. All results are based on the DIV2K validation set.}
    \label{fig:curves}
    \vspace{-10pt}
\end{figure}

\begin{figure*}
    \centering
    \includegraphics[width=0.95\linewidth]{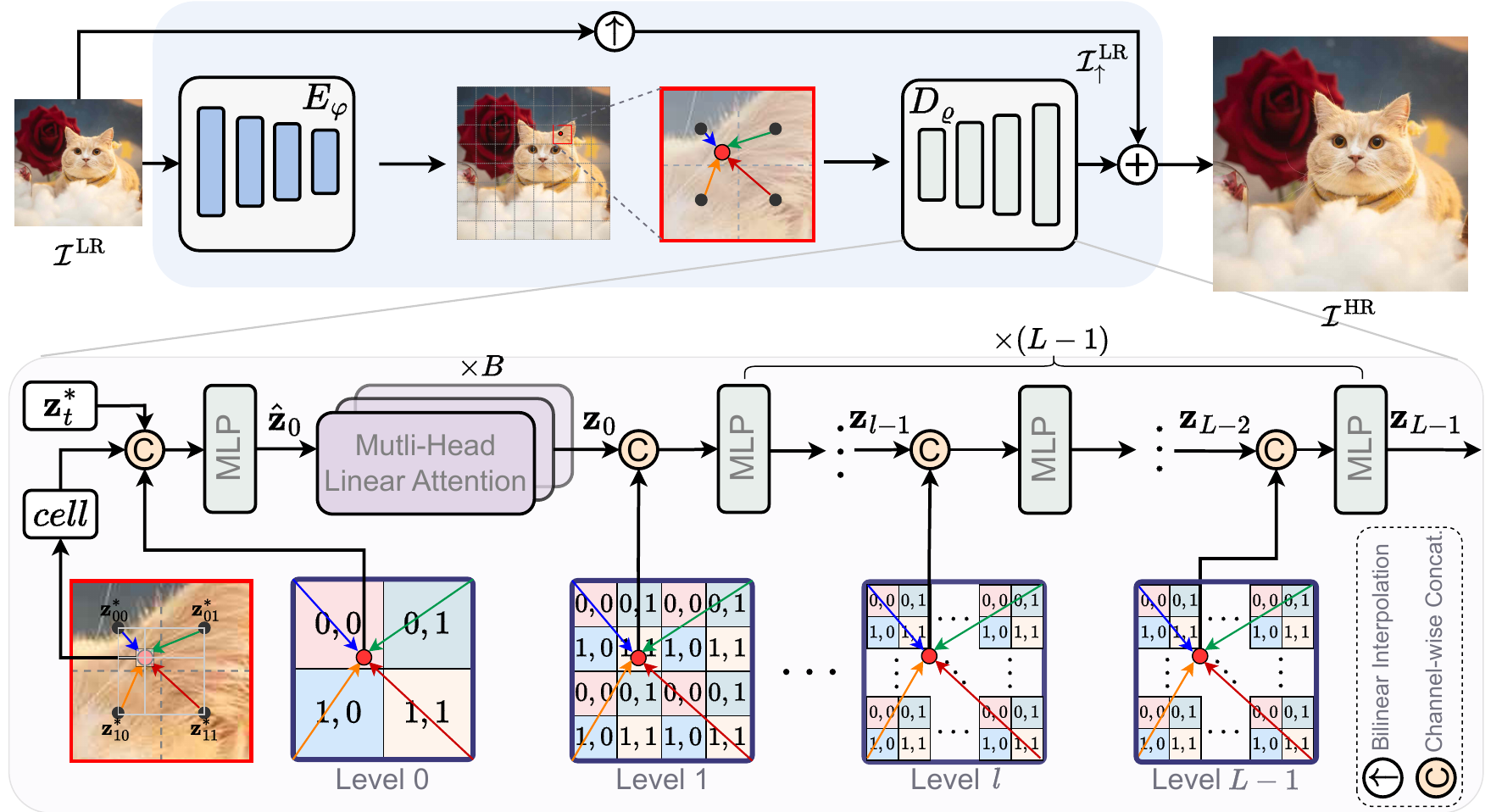}
    \caption{Illustration of the proposed HIIF framework. The encoder here can be replaced by any existing ISR architectures.}
    \label{fig:HIIFarchi}
\end{figure*}

Recent work on arbitrary-scale ISR methods has often exploited implicit neural representations (INRs) \cite{chen2021learning, lee2022local,yao2023local}. INRs were first proposed to build a bridge between discrete and continuous representation, optimized to efficiently model complex data types such as images \cite{sitzmann2020implicit}, videos \cite{chen2021nerv, kwan2023hinerv}, and 3D scenes \cite{chibane2020implicit, nerfstudio, mildenhall2021nerf}. Unlike other methods that rely on discrete grids or pixel-based formats, INRs use coordinate-based neural networks, typically multilayer perceptrons (MLPs), to map input coordinates to signal values \cite{chen2021nerv}. This enables INRs to reconstruct fine details with fewer parameters, offering a flexible and efficient solution for high-dimensional data representation. When used for continuous super-resolution applications, INRs have demonstrated great potential compared to other learning-based methods. Notable works include LIIF \cite{chen2021learning} that achieves a mapping across spatial dimensions and LTE \cite{lee2022local} which further enhances textures in the frequency domain. More recent work has focused on reconstructing sharper details, with CiaoSR \cite{cao2023ciaosr} and CLIT \cite{chen2023cascaded} as typical examples. Despite these advances, existing methods often represent data at a single scale. In contrast, multi-scale representations have demonstrated superior ability for different image processing and computer vision applications \cite{dong2015image, mildenhall2021nerf, chen2022videoinr, jiao2021multiscale, he2019dynamic}.

In the above context, this paper proposes a \textbf{H}ierarchical encoding based \textbf{I}mplicit \textbf{I}mage \textbf{F}unction for continuous super-resolution, \textbf{HIIF}. By encoding relative positional information hierarchically, HIIF implicitly represents local features at multiple scales, enhancing the connections between sampling points within local regions which leads to improved representational ability. Moreover, we incorporate a multi-head self-attention mechanism within our representation network to expand the receptive field and fully leverage non-local information. Unlike previous approaches \cite{chen2021learning, lee2022local}, HIIF better exploits spatial information in latent space to address the limitations of linear interpolation in capturing high-frequency details. 
The main contributions of this work are summarized as follows:

\begin{itemize}
    \item \textbf{A novel implicit representation framework} for continuous image super-resolution based on a new \textbf{hierarchical position encoding} network. This is the first time that multi-scale hierarchical position encoding has been used for super-resolution. Existing works \cite{chen2021learning, lee2022local, chen2023cascaded} are typically based on single-scale position encoding.
    \item \textbf{A new multi-scale architecture} to concatenate the local features with relative coordinates and implicitly learn to aggregate the output. This is different from the local ensemble methods in existing continuous super-resolution models \cite{chen2021learning, lee2022local, cao2023ciaosr}, which are based on fixed or learnable ensemble weights.
    \item The employment of a \textbf{multi-head linear attention module} to enhance the model's ability to capture information in different representation subspaces. This is also the first time this type of approach has been used for super-resolution. 
\end{itemize}

We have evaluated the proposed framework by integrating it with three commonly used backbone encoders, EDSR \cite{lim2017enhanced}, RDN \cite{zhang2018residual} and SwinIR \cite{liang2021swinir}. The results (summarized in Figure \ref{fig:curves}) show that, when compared to existing continuous super-resolution models, HIIF achieves consistent PSNR gains across different in-distribution and out-of-distribution up-sampling ratios (up to 0.17dB).  HIIF is a flexible framework that can be seamlessly integrated with different backbone encoder networks (for feature extraction).

%% file: sec/2_related_work.tex
\section{Related work}
\label{sec:RW}

\textbf{Image super-resolution} (ISR) has attracted increased research interest over the past few decades. Its aim is to generate a high-resolution image from a low-resolution version, with the goal of achieving better perceptual quality while accurately recovering spatial details. Leveraging the advances in deep learning, numerous architectures have emerged since the introduction of the first important ISR model, SRCNN \cite{dong2015image}. These learning-based approaches can be categorized into four primary classes according to their architecture: CNN-based \cite{dong2015image, kim2016accurate, lim2017enhanced, zhang2018image}; transformer-based \cite{liang2022vrt, wang2022uformer, liang2021swinir, conde2022swin2sr}; diffusion-based \cite{gao2023implicit, saharia2022image}; and SSM-based methods \cite{guo2024mambair, ren2024mambacsr, shi2024vmambair}. EDSR \cite{lim2017enhanced}, SwinIR \cite{liang2021swinir}, IDM \cite{gao2023implicit} and MambaIR \cite{guo2024mambair} are typical examples in each category, respectively.

\noindent\textbf{Implicit neural representations} (INRs) have gained popularity in recent years due to their performance on low-level vision tasks such as 3D view synthesis \cite{nerfstudio}, object shape modeling \cite{atzmon2020sal}, structure rendering \cite{chabra2020deep, chen2019learning, nerfstudio}, image \cite{klocek2019hypernetwork, sitzmann2020implicit, chen2021learning, wu2023neural}/video \cite{chen2021nerv, kwan2023hinerv, kwan2024nvrc, gao2024pnvc} representation and compression. In these applications, most research contributions have utilized coordinate-based multilayer perceptrons (MLPs) to represent continuous-domain signals by mapping the coordinates to target values, such as from pixel coordinates to image RGB color values. While many of these optimize a single representation for a signal instance, a class of them \cite{ma2024implicit, chen2021learning} learn an implicit neural representation from a dataset and utilize them as a generalized function that can be used for different data instances. Typically, a hypernetwork encodes prior information in a latent space \cite{chauhan2024brief}, allowing the representation to be data-dependent and facilitating knowledge sharing across diverse samples. 

%-------------------------------------------------------------------------
\noindent\textbf{Arbitrary-scale super-resolution.} Most ISR techniques focus on fixed up-sampling factors (e.g., $\times$2, $\times$3, $\times$4) and have achieved impressive results \cite{lim2017enhanced, liang2021swinir, zhang2018residual, zhang2018image}. However, these approaches lack flexibility, usually requiring different models for various scaling factors. Recently, several learning-based methods, such as MetaSR \cite{hu2019meta}, have been developed to perform arbitrary-scale super-resolution with a single model. Inspired by the implicit neural representation approaches \cite{park2019deepsdf, mescheder2019occupancy, sitzmann2020implicit}, LIIF \cite{chen2021learning} predicts the signal for arbitrary coordinates by learning implicit features from local regions, delivering promising results for both \textit{in-distribution} and \textit{out-of-distribution} scaling factors. Subsequently, LTE \cite{lee2022local} was developed, focusing on estimating dominant frequencies and corresponding Fourier coefficients transformed from the latent feature space in the frequency domain. To further improve local ensembling, CiaoSR \cite{cao2023ciaosr} explicitly learns ensemble weights and leverages scale-aware information. More recently, SRNO \cite{wei2023super} employs a super-resolution neural operator that maps between finite-dimensional function spaces, while CLIT \cite{chen2023cascaded} incorporates a local attention mechanism, a cumulative training strategy and a cascaded framework to achieve large-scale up-sampling. It is noted that, although these INR-based methods have delivered state-of-the-art results for the arbitrary-scale super-resolution, they rarely consider the connection between local features, simply adopting a single-scale representation. To address these issues, we investigate the use of a multi-scale representation for continuous super-resolution. 

%% file: sec/3_method.tex
\section{Method}
\label{sec:Method}

\subsection{Problem formulation}

%-------------------------------------------------------------------------
% \subsection{Local implicit neural representation}
Based on the Local Implicit Image Function (LIIF) \cite{chen2021learning}, a decoding function $f_{\theta}$ is used to map from a 2D coordinate $\mathbf{c}=[x,y]\in  \mathbb{R}^{2}$ to RGB values $\mathbf{s}=[R,G,B] \in  \mathbb{R}^{3}$:
 \begin{equation}
\mathbf{s} = f_{\theta}(\mathbf{z},\mathbf{c}, cell),     
 \end{equation}
where $\theta$ represents the trainable parameters of $f$. $cell = [\frac{2}{r_yH}, \frac{2}{r_xW}]$ represents the cell decoding, with variable scaling factors $r_{x}$ and $r_{y}$ and the original image shape $H$ and $W$. $\mathbf{z}$ is the latent code, extracted by an encoder $E_{\varphi}$ from a given low-resolution image $\mathcal{I}^\mathrm{LR} \in \mathbb{R}^{H \times W \times 3}$, 
\begin{equation}
\mathbf{z} = E_{\varphi}(\mathcal{I}^\mathrm{LR}).   
 \end{equation}
The implicit image function is typically parameterized by MLPs, trained on a large dataset, and applied to any image during inference. In order to recover the high resolution version of the image, $\mathcal{I}^\mathrm{HR} \in \mathbb{R}^{r_{y}H \times r_{x}W \times 3}$, the RGB values at the queried coordinate $\mathbf{x}_{q} \in \mathbb{R}^{2}$ is reconstructed according to the surrounding nearest (based on the Euclidean distance) latent codes $\{\mathbf{z}_t^{*}\}$, $t \in \{00, 01, 10, 11\}$,
 \begin{equation}
\mathcal{I}^\mathrm{HR}(\mathbf{x}_{q}) = \sum_{t} w_{t} f_{\theta}(\mathbf{z}_t^{*}, \delta (\mathbf{x}_{q}, t), cell), 
 \end{equation}
\begin{equation}
\delta (\mathbf{x}_{q}, t) = \mathbf{x}_{q} - \mathbf{x}_t^{*},
\end{equation}
where $\mathbf{x}_t^{*}$ is the corresponding coordinate of the nearest latent code (for $t$), and $\delta (\mathbf{x}_{q}, t)$ represents the relative coordinate, which is also known as the local grid. Based on the fixed ensemble weight method proposed in \cite{chen2021learning}, $w_{t}$ is calculated according to the area of the rectangle formed by $\mathbf{x}_{q}$ and $\mathbf{x}_t^{*}$, and normalized to satisfy $\sum_{t} w_{t} = 1$.  
%-------------------------------------------------------------------------

\subsection{Overall design}
As illustrated in Figure \ref{fig:HIIFarchi}, the proposed continuous super-resolution framework with the hierarchical encoding based implicit image function, HIIF, utilizes a latent encoder $E_{\varphi}$ to extract latent features from the input low-resolution image, $\mathcal{I}^\mathrm{LR}$. With the extracted latent features, for each queried coordinate, its four nearest latent codes are then identified and fed into the decoder $D_{\varrho}$.  A skip connection is employed to connect the bilinearly up-sampled \cite{lee2022local} input image, $\mathcal{I}^\mathrm{LR}_{\uparrow}$, and the output of $D_{\varrho}$ to produce the final high-resolution image, $\mathcal{I}^\mathrm{HR}$. The complete workflow can be described by the following equation:
\begin{align}
\mathcal{I}^\mathrm{HR}(\mathbf{x}_q) &=D_{\varrho}\left(E_{\varphi}(\mathcal{I}^\mathrm{LR}), \{ \delta_{h} (\mathbf{x}_{q}, l) \}, cell\right) + \mathcal{I}^\mathrm{LR}_{\uparrow}(\mathbf{x}_q).
\end{align}
% \begin{align}
% \mathcal{I}^\mathrm{HR}(\mathbf{x}_q) &= \mathcal{I}^\mathrm{LR}_{D}(\mathbf{x}_q) + \mathcal{I}^\mathrm{LR}_{\uparrow}(\mathbf{x}_q), \\
% \mathcal{I}^\mathrm{LR}_{D}(\mathbf{x}_q) &=  D_{\varphi}\left(E_{\varphi}(\mathcal{I}^\mathrm{LR}), \delta_{h} (\mathbf{x}_{q}, 0), ..., \delta_{h} (\mathbf{x}_{q}, L - 1), cell\right).
% \end{align}
%
Here $\delta_{h} (\mathbf{x}_{q}, l)$ denotes the $l$-th level hierarchical encoding \cite{kwan2023hinerv}, in which $l \in \{0, 1, \ldots, L-1\}$.

\subsection{Encoder}
Following previous work \cite{chen2021learning, lee2022local} on continuous super-resolution, the encoder $E_{\varphi}$ in our framework is employed to generate the latent codes from the input: $\mathbb{R}^{H \times W \times 3} \mapsto \mathbb{R}^{H \times W \times C_{enc}}$, $C_{enc}$ is the channel dimension of the encoder. As there is no down/up-sampling layer within the encoder, the output has the same spatial size as the input $\mathcal{I}^\mathrm{LR}$. Afterward, a convolutional layer is connected: $\mathbb{R}^{H \times W \times  C_{enc}} \mapsto \mathbb{R}^{H \times W \times C}$, i.e., $\mathbf{z}_t^{*} \in \mathbb{R}^{H \times W \times C}$. In this paper, we follow existing works \cite{chen2021learning, lee2022local} to use EDSR \cite{lim2017enhanced}, RDN \cite{zhang2018residual}, and SwinIR \cite{liang2021swinir} as encoder backbones to validate our proposed method.

\begin{table*}[!t]
\centering
\caption{Quantitative comparison results on the DIV2K \cite{timofte2017ntire} validation set and Set5 \cite{bevilacqua2012low} dataset. For each column, the best result is colored in \textcolor{red}{red} and the second best is colored in \textcolor{blue}{blue}. `-' indicates that the result is not available in the literature (or the source code of the model has not been released).}
\resizebox{\linewidth}{!}{\begin{tabular}{c|r|ccc|ccccc|ccc|ccc}
\toprule
\multirow{3}{*}{\rotatebox{270}{Encoder}}&\multicolumn{1}{c|}{Database}& \multicolumn{8}{c|}{DIV2K} & \multicolumn{6}{c}{Set5}\\
\cmidrule{2-16}
&\multicolumn{1}{c|}{PSNR (dB)$\uparrow$} & \multicolumn{3}{c|}{In-distribution} & \multicolumn{5}{c|}{Out-of-distribution} & \multicolumn{3}{c|}{In-distribution} & \multicolumn{3}{c}{Out-of-distribution} \\ \cmidrule{2-16}
 & Method & $\times 2$ & $\times 3$ & $\times 4$ & $\times 6$ & $\times 12$ & $\times 18$ & $\times 24$ & $\times 30$ & $\times 2$ & $\times 3$ & $\times 4$ & $\times 6$ & $\times 8$ & $\times 12$\\ \midrule
n/a   & \textit{Bicubic} & 31.01 & 28.22 & 26.66 & 24.82 & 22.27 & 21.00 & 20.19 & 19.59 & 32.30 & 29.04 & 27.06 & 24.58 & 23.00 & 21.24 \\ \midrule
\multirow{8}{*}{\rotatebox{270}{EDSR\_baseline \cite{lim2017enhanced}}} & EDSR only & 34.52 & 30.89 & 28.98 & - & - & - & - & - & 37.99 & 34.36 & 32.09 & - & - &-\\
 \cmidrule{2-16}
     & MetaSR \cite{hu2019meta} & 34.64 & 30.93 & 28.92 & 26.61 & 23.55 & 22.03 & 21.06 & 20.37 & 37.94 & 34.35 & 32.07 & 28.74 & 26.83 & 24.53   \\
     & LIIF \cite{chen2021learning} & 34.67 & 30.96 & 29.00 & 26.75 & 23.71 & 22.17 & 21.28 & 20.48 &  37.97 & 34.39 & 32.22 & 28.93 & 26.98 & 24.57 \\ 
     & LTE \cite{lee2022local} & 34.72 & 31.02 & 29.04 & 26.81 & 23.78 & 22.23 & 21.24 & 20.53  & 38.02 & 34.42 & 32.22 & 28.95 & 27.02 & 24.60   \\ 
     & CLIT \cite{chen2023cascaded} & 34.81 & \textcolor{blue}{31.12} & 29.15 & \textcolor{blue}{26.92} & 23.83 & 22.29 & 21.26  & 20.53  &- &- &- &- &- &-   \\ 
     & CiaoSR \cite{cao2023ciaosr} & \textcolor{blue}{34.88} & \textcolor{blue}{31.12} & \textcolor{blue}{29.19} & \textcolor{blue}{26.92} & \textcolor{blue}{23.85} & \textcolor{blue}{22.30} & \textcolor{red}{21.29} & 20.44   & \textcolor{red}{38.13} & 34.47 & \textcolor{red}{32.42} & \textcolor{red}{29.10} & \textcolor{red}{27.12} & \textcolor{blue}{24.68}  \\ 
     & SRNO \cite{wei2023super} & 34.85 & 31.11 & 29.16 & 26.90 & 23.84 & 22.29 & 21.27 & \textcolor{red}{20.56}  & \textcolor{blue}{38.11} & \textcolor{blue}{34.51} & \textcolor{blue}{32.37} & 29.02 & 27.03 & 24.60   \\ 
      \cmidrule{2-16}
    & \textbf{HIIF (ours)} & \textcolor{red}{34.90} & \textcolor{red}{31.19} & \textcolor{red}{29.21} & \textcolor{red}{26.94} & \textcolor{red}{23.87} & \textcolor{red}{22.31} & \textcolor{blue}{21.27} & \textcolor{blue}{20.54}  & \textcolor{blue}{38.11} & \textcolor{red}{34.54} & \textcolor{red}{32.42} & \textcolor{blue}{29.06} & \textcolor{blue}{27.08} & \textcolor{red}{24.69}  \\ \midrule

\multirow{8}{*}{\rotatebox{270}{RDN\cite{zhang2018residual}}} & RDN only & 34.59 & 31.03 & 29.12 & - & - & - & - & -  &38.24  &34.71  &32.47 & -& - &-\\ 
 \cmidrule{2-16}
     & MetaSR \cite{hu2019meta} & 35.00 & 31.27 & 29.25 & 26.88 & 23.73 & 22.18 & 21.17 & 20.47 & 38.22  &34.63  &32.38 & 29.04 &26.96& - \\ 
     & LIIF \cite{chen2021learning} & 34.99 & 31.26 & 29.27 & 26.99 & 23.89 & 22.34 & 21.31 & 20.59  &38.17 & 34.68  &32.50 &29.15& 27.14 &24.86\\ 
     & LTE \cite{lee2022local} & 35.04 & 31.32 & 29.33 & 27.04 & 23.95 & 22.40 & 21.36 & \textcolor{blue}{20.64} & 38.23 & 34.72  &32.61& 29.32& 27.26 &24.79 \\ 
     & CLIT \cite{chen2023cascaded} & 35.10 & 31.39 & 29.39 & 27.12 & 24.01 & 22.45 & 21.38 & \textcolor{blue}{20.64}  &38.26  &34.79 & \textcolor{blue}{32.69} &\textcolor{red}{29.54}& 27.34& -   \\ 
     & CiaoSR \cite{cao2023ciaosr} & 35.13 & 31.39 & \textcolor{blue}{29.43} & \textcolor{blue}{27.13} & \textcolor{red}{24.03} & 22.45 & \textcolor{red}{21.41} & 20.55 & 38.29 & \textcolor{red}{34.85} & 32.66&  29.46 &\textcolor{red}{27.36} &\textcolor{blue}{24.92} \\ 
     & SRNO \cite{wei2023super} & \textcolor{blue}{35.16} & \textcolor{blue}{31.42} & 29.42 & 27.12 & \textcolor{red}{24.03} & \textcolor{blue}{22.46} & \textcolor{red}{21.41} & \textcolor{red}{20.68}  &\textcolor{blue}{38.32}  &\textcolor{blue}{34.84} & \textcolor{blue}{32.69} &29.38 &27.28& -  \\ 
      \cmidrule{2-16}
    & \textbf{HIIF (ours)} & \textcolor{red}{35.27} & \textcolor{red}{31.59} & \textcolor{red}{29.56} & \textcolor{red}{27.16} & \textcolor{blue}{24.02} & \textcolor{red}{22.48} & \textcolor{blue}{21.40} & 20.62 & \textcolor{red}{38.34} & \textcolor{red}{34.85} & \textcolor{red}{32.70} & \textcolor{blue}{29.49} & \textcolor{blue}{27.35} & \textcolor{red}{24.95} \\ \midrule

\multirow{8}{*}{\rotatebox{270}{SwinIR\cite{liang2021swinir}}} & SwinIR only & 35.09 & 31.42 & 29.48 & - & - & - & - & - &38.35& 34.89 &32.72 &-&-&-\\  
\cmidrule{2-16}
     & MetaSR \cite{hu2019meta} & 35.15 & 31.40 & 29.33 & 26.94 & 23.80 & 22.26 & 21.26 & 20.54 &38.26 &34.77& 32.47  &29.09 &27.02& 24.82 \\ 
     & LIIF \cite{chen2021learning} & 35.17 & 31.46 & 29.46 & 27.15 & 24.02 & 22.43 & 21.40 & 20.67& 38.28& 34.87 &32.73& 29.46 &27.36 & 24.98   \\ 
     & LTE \cite{lee2022local} & 35.24 & 31.50 & 29.51 & 27.20 & 24.09 & 22.50 & 21.47 & \textcolor{blue}{20.73} &38.33& 34.89& 32.81 &29.50& 27.35& \textcolor{blue}{25.01}  \\ 
     & CLIT \cite{chen2023cascaded} & \textcolor{blue}{35.29} & \textcolor{blue}{31.55} & 29.55 & \textcolor{blue}{27.26} & 24.11 & \textcolor{blue}{22.51} & 21.45 & 20.70 &\textcolor{blue}{38.41}& \textcolor{blue}{34.97}& \textcolor{blue}{32.86} & \textcolor{red}{29.69} &\textcolor{red}{27.62}& - \\ 
     & CiaoSR \cite{cao2023ciaosr} & 35.28 & 31.53 & \textcolor{blue}{29.56} & \textcolor{blue}{27.26} & \textcolor{blue}{24.12} &\textcolor{red}{22.52} & \textcolor{red}{21.48} & 20.59 &38.38 &34.91& 32.84 &29.62& 27.45& 24.96 \\ 
     \cmidrule{2-16}
    & \textbf{HIIF (ours)} & \textcolor{red}{35.34} & \textcolor{red}{31.62} & \textcolor{red}{29.61} & \textcolor{red}{27.27} & \textcolor{red}{24.13} & 22.50 & \textcolor{red}{21.48} & \textcolor{red}{20.73} & \textcolor{red}{38.43} & \textcolor{red}{34.98} & \textcolor{red}{32.87} & \textcolor{blue}{29.64} & \textcolor{blue}{27.56} & \textcolor{red}{25.03}   \\ \bottomrule
    % & \textbf{HIIF (ours)} & \textcolor{red}{35.45} & \textcolor{red}{31.73} & \textcolor{red}{29.71} & \textcolor{red}{27.30} & \textcolor{red}{24.14} & \textcolor{red}{22.54} & \textcolor{red}{21.49} & \textcolor{red}{20.75} & \textcolor{red}{38.43} & \textcolor{red}{34.98} & \textcolor{red}{32.87} & \textcolor{blue}{29.64} & \textcolor{blue}{27.56} & \textcolor{red}{25.03}   \\ \bottomrule
% Add additional model rows here
\end{tabular}}
\label{tbl:results1}
\vspace{-5pt}
\end{table*}

\subsection{Decoder}

As illustrated in Figure \ref{fig:HIIFarchi}, the proposed HIIF decoder consists of a multi-scale hierarchical encoding module, $B$ multi-head linear attention blocks and multiple MLPs. Their detailed designs are described as follows.
%-------------------------------------------------------------------------
\subsubsection{Hierarchical encoding}

\label{subsection:he}
\begin{figure}[!t]
    \centering
    \includegraphics[width=0.95\linewidth]{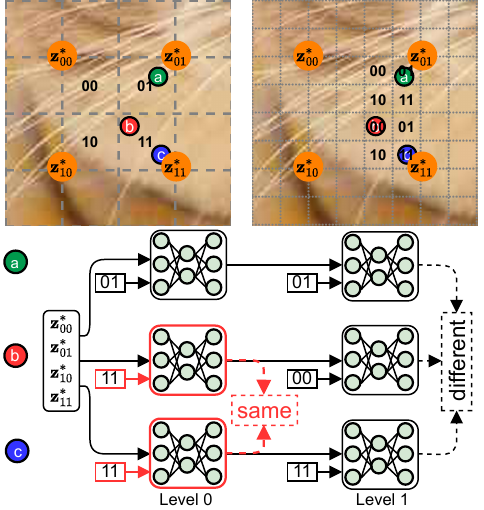}
    \caption{The proposed multi-scale architecture. By applying hierarchical encoding at different levels in the decoder, the sampling points share the same network features with the neighboring points at the coarser levels but not at the finer levels. 
    %In the example above, at the first level, all points share the same hierarchical encoding, which is passed to the decoder layer, resulting in identical outputs for each point. However, at the second level, only points B and C share the same encoding, and therefore, the output from the decoder layer is the same for these two points. At level 3, the outputs for A, B and C are all different.}
    }
    \label{fig:hidraft}
\end{figure}

% It is noted that, although neighboring features exhibit strong correlations, parameter-free interpolation methods often fail to accurately capture high-frequency details compared to convolutional layers. One solution to this problem is to introduce positional encoding during up-sampling. 
Existing methods based on implicit neural representations \cite{lee2022local, lee2022local, chen2023cascaded, cao2023ciaosr, wei2023super} use positional input to enable arbitrary-scale decoding. Although the use of the relative positional input provides information required for the local implicit image function to decode (reconstruct) the target output \cite{chen2021learning, chen2023cascaded}, such neural representation models do not fully consider the neighborhood relationships between local features, because these MLP-based models are not explicitly designed for multi-scale representation. 

Inspired by \cite{kwan2023hinerv, li2024asmr} that utilizes hierarchical-based positional encoding or modulation for implicit neural representations, we propose a novel local hierarchal positional encoding (shown in Figure \ref{fig:HIIFarchi}), which explicitly models the local implicit image function as a multi-scale representation. In our method, the original relative coordinate is reparameterized into a set of encodings at different scales. By conditioning the neural network layers with these encoding sequentially, the intermediate features at different scales are effectively shared by neighboring sampling points, as illustrated in Figure \ref{fig:hidraft}.

Specifically, for each coordinate $(x, y)$, the local coordinate $(x_{local}, y_{local})$, which is relative to the position of feature $\mathbf{z}^{*}_{00}$, is first calculated and normalized to $[0, 1]$. The hierarchical coordinate is then obtained by
\begin{equation}
    (x_{hier}, y_{hier}) = \lfloor (x_{local}, y_{local}) \times S^{l + 1}\rfloor \bmod S,
\end{equation}
where $S$ is the scaling factor. $(x_{hier}, y_{hier})$ is embedded as the hierarchical encoding $\delta_{h} (\mathbf{x}_{q}, l)$.

\subsubsection{Multi-scale architecture}
\label{subsection:mg}

Although hierarchical encodings provide positional information at different scales, directly applying them to the decoder does not ensure the learning of multi-scale features. In this work, inspired by \cite{wu2023neural, kwan2023hinerv, li2024asmr}, instead of applying hierarchical encodings directly to the decoder, these encodings are progressively incorporated into different network layers. This approach exploits the connection of neighboring features, and allows each level of network layers to focus on a specific frequency subband or level of detail, as shown in Figure \ref{fig:HIIFarchi}. Specifically, the feature output at the $0$-th level, denoted as $\mathbf{z}_0$, $\mathbf{z}_0 \in \mathbb{R}^{H \times W \times C}$, is calculated by
\begin{align}
 \mathbf{\hat{z}_{0}} &= \mathrm{MLP}([\{ \mathbf{z^{*}_t}\}, \{\delta_{h} (\mathbf{x}_{q}, 0)\}, cell]),\\
 \mathbf{z_{0}} &= \mathrm{MHA}([\mathbf{\hat{z}}_0]).
\end{align}
where $t \in \{00, 01, 10, 11\}$. $\mathrm{MHA}$ denotes the multi-head attention layer (implementation details are provided below). The input of the other levels ($l>1$) is $\mathbf{z}_{l-1}$, guided by the $l$-th level hierarchical encoding, $\delta_{h} (\mathbf{x}_{q}, l)$. %MHA is only applied to early levels.

% %
% \begin{align}
%  \mathbf{z}_l &= \mathrm{MLP}([\mathbf{z}_{l-1}, \{\delta_{h} (\mathbf{x}_{q}, l)\}])
% \end{align}
% %
%where $l \in \{0, ..., L-1\}$. 
%After the $L-1$-th level, a linear layer is used to predict the final output color.

%-------------------------------------------------------------------------
\subsubsection{Multi-head linear attention}
\label{subsection:mhla}

As reported in \cite{cao2023ciaosr}, the use of attention within the implicit function is non-trivial because standard self-attention and neighborhood attention mechanisms do not incorporate coordinate information as a conditioning factor. Multi-head attention \cite{liu2021swin,liang2021swinir} is a mechanism in deep learning used to enhance the ability of a model to capture information in different representation subspaces. It helps to learn complex patterns by simultaneously attending to different parts of the input using multiple independent attention heads. In this work, we have employed a multi-head linear attention block based on \cite{ali2021xcit, shen2021efficient} in order to maintain a relatively low computational complexity.
%then split it into $N$ $\hat{\mathbf{z}}_0^{n} \in \mathbb{R}^{N \times HW \times \frac{C}{N}}$, where
Specifically, the input $\hat{\mathbf{z}}_0$ is firstly reshaped to $\mathbb{R}^{HW \times C}$, and $N$ heads are employed. Three learnable projection matrices $\mathbf{W}^{n}_Q, \mathbf{W}^{n}_K, \mathbf{W}^{n}_V \in \mathbb{R}^{C \times \frac{C}{N}}$ are then used to obtain queries \( \mathbf{Q}_n \), keys \( \mathbf{K}_n \), and values $ \mathbf{V}_n \in \mathbb{R}^{HW \times  \frac{C}{N} }$:
\begin{equation}
\mathbf{Q}_n = \hat{\mathbf{z}}_0 \mathbf{W}^{n}_Q, \quad \mathbf{K}_n = \hat{\mathbf{z}}_0 \mathbf{W}^{n}_K, \quad \mathbf{V}_n = \hat{\mathbf{z}}_0 \mathbf{W}^{n}_V.
\end{equation}
Here, layer normalization is applied to $\mathbf{K}_n$ and $\mathbf{Q}_n$. Based on these, the attention is calculated by,
\begin{equation}
\text{Attention}(\mathbf{Q}_n, \mathbf{K}_n, \mathbf{V}_n) = \mathbf{V}_n \times (\frac{\mathbf{K}_n^T \mathbf{Q}_n}{\sqrt{HW}}).
\end{equation}
Finally, the reformatted $\text{Attention}(\mathbf{Q}_n, \mathbf{K}_n, \mathbf{V}_n)$ is reshaped back to its original size, combined with the input of this multi-head attention block through skip connection.

%% file: sec/4_Experiment.tex
\section{Experiment}
\label{sec:experiment}

% In this section, we present the experimental results and discuss their implications. We begin with a brief introduction to our experimental setup in Section 4.1. Following that, our proposed HIIF is evaluated and compared with other methods based on different datasets in Section 4.2. Then, Section 4.3 compares the qualitative results between various encoders with our HIIF. Finally, ablation studies for various configurations of the proposed HIIF are compared in Section 4.4.

\begin{table*}[!t]
\centering
\caption{Quantitative comparison results on the Set14 \cite{zeyde2012single}, BSD100 \cite{martin2001database}, and Urban100 \cite{huang2015single} datasets in terms of PSNR. For each column, the best result is colored in \textcolor{red}{red} and the second best is colored in \textcolor{blue}{blue}. `-' indicates that the result is not available in the literature (or the source code of the model has not been released).}
\resizebox{\linewidth}{!}{\begin{tabular}{c|r|ccc|ccc|ccc|ccc|ccc|ccc}
\toprule
\multirow{3}{*}{\rotatebox{270}{Encoder}}&\multicolumn{1}{c|}{Database}& \multicolumn{6}{c|}{Set14} & \multicolumn{6}{c|}{BSD100} &\multicolumn{6}{c}{Urban100}\\
\cmidrule{2-20}
 & \multicolumn{1}{c|}{PSNR (dB)$\uparrow$} & \multicolumn{3}{c|}{In-distribution} & \multicolumn{3}{c|}{Out-of-distribution} & \multicolumn{3}{c|}{In-distribution} & \multicolumn{3}{c|}{Out-of-distribution}& \multicolumn{3}{c|}{In-distribution} & \multicolumn{3}{c}{Out-of-distribution} \\ \cmidrule{2-20}
 & Method & $\times 2$ & $\times 3$ & $\times 4$ & $\times 6$ & $\times 8$ & $\times 12$ & $\times 2$ & $\times 3$ & $\times 4$ & $\times 6$ & $\times 8$ & $\times 12$& $\times 2$ & $\times 3$ & $\times 4$ & $\times 6$ & $\times 8$ & $\times 12$ \\ \midrule
 n/a & \textit{Bicubic} & 28.72 & 26.04 & 24.50 & 22.74 & 21.64 & 20.27 & 28.24 & 25.89 & 24.65 & 23.25 & 22.40 & 21.32 & 25.54 & 23.12 & 21.81 & 20.31 & 19.42 & 18.33 \\ \midrule
\multirow{7}{*}{\rotatebox{270}{EDSR\_baseline \cite{lim2017enhanced}}} & EDSR only & 33.56 & 30.28 & 28.56 & - & - & - & 32.17 & 29.09 & 27.57 & - & - & - & 31.99 & 28.15 & 26.03 & - & - & -  \\ \cmidrule{2-20}
     & MetaSR \cite{hu2019meta} & 33.58 & 30.31 & 28.52 & 26.38 & 24.85 & 23.13 & 32.19 & 29.10 & 27.56 & 25.79 & 24.74 & 23.45 & 32.10 & 28.15 & 26.05 & 23.69 & 22.37 & 20.85 \\ \cmidrule{2-20}
     & LIIF \cite{chen2021learning} & 33.66 & 30.33 & 28.60 & 26.44 & 24.93 & 23.19 & 32.19 & 29.11 & 27.60 & 25.84 & 24.78 & 23.49 & 32.17 & 28.22 & 26.14 & 23.78 & 22.44 & 20.91  \\  \cmidrule{2-20}
     & LTE \cite{lee2022local} & 33.70 & 30.36 & 28.63 & 26.48 & 24.97 & 23.20 & 32.22 & 29.15 & 27.62 & 25.87 & 24.81 & 23.50 & 32.30 & 28.32 & 26.23 & 23.84 & 22.52 & 20.96  \\  \cmidrule{2-20}
     % & CLIT \cite{chen2023cascaded} &  &  &  & - & - & - & - & - & & & & & & & & & &  \\ \cmidrule{2-20}
     & CiaoSR \cite{cao2023ciaosr} & \textcolor{blue}{33.88}  & 30.46 & 28.76 & \textcolor{red}{26.59} & \textcolor{red}{25.07} & \textcolor{blue}{23.23} & \textcolor{blue}{32.28} & \textcolor{blue}{29.20} & \textcolor{blue}{27.70} & \textcolor{red}{25.94} & \textcolor{blue}{24.88} & \textcolor{blue}{23.55} & \textcolor{red}{32.79} & \textcolor{blue}{28.57} & \textcolor{red}{26.58} & \textcolor{red}{24.23} & \textcolor{red}{22.82} & \textcolor{blue}{21.08} \\ \cmidrule{2-20}
     & SRNO \cite{wei2023super} & 33.82 & \textcolor{blue}{30.49} & \textcolor{blue}{28.77} & \textcolor{blue}{26.53} & \textcolor{blue}{25.02} & 23.16 & \textcolor{blue}{32.28} & 29.19 & 27.67 & 25.90 & 24.86 & 23.52 & 32.63 & \textcolor{blue}{28.57} &  26.49 & \textcolor{blue}{24.06} & 22.66 & 21.07 \\ \cmidrule{2-20}
    & \textbf{HIIF (ours)} & \textcolor{red}{33.88} & \textcolor{red}{30.51} & \textcolor{red}{28.81} & \textcolor{red}{26.59} & \textcolor{blue}{25.02} & \textcolor{red}{23.24} & \textcolor{red}{32.33} & \textcolor{red}{29.25} & \textcolor{red}{27.71} & \textcolor{blue}{25.93} & \textcolor{red}{24.89} & \textcolor{red}{23.56} & \textcolor{blue}{32.69} & \textcolor{red}{28.59} & \textcolor{blue}{26.51} & 23.99 & \textcolor{blue}{22.72} & \textcolor{red}{21.09} \\ \midrule
\multirow{8}{*}{\rotatebox{270}{RDN\cite{zhang2018residual}}} & RDN only & 34.01 & 30.57 & 28.81  & - & - & - & 32.34& 29.26 &27.72 & -&  -&  -& 32.89& 28.80 &26.61& -&  -&  - \\ \cmidrule{2-20}
     & MetaSR \cite{hu2019meta} & 33.98 & 30.54 & 28.78  & 26.51& 24.97 &-  & 32.33 &29.26 &27.71  &25.90& 24.97 &-& 32.92 &28.82& 26.55& 23.99& 22.59& -\\ 
     & LIIF \cite{chen2021learning} & 33.97& 30.53& 28.80 & 26.64& 25.15 &23.24  & 32.32 &29.26& 27.74 & 25.98 & 24.91 & 23.57&  32.87& 28.82 &26.68& 24.20& 22.79& 21.15 \\ 
     & LTE \cite{lee2022local} & 34.09 &30.58& 28.88 & 26.71& 25.16 &23.31  & 32.36 &  29.30  & 27.77 & 26.01&  24.95&  23.60 &33.04& 28.97& 26.81 & 24.28& 22.88& 21.22 \\ 
     & CLIT \cite{chen2023cascaded} & 34.09& 30.69& \textcolor{blue}{28.93} & \textcolor{blue}{26.83} & \textcolor{red}{25.36} & -  & 32.39& 29.33& 27.80 & \textcolor{blue}{26.07} & \textcolor{blue}{25.00} &  - &33.14 &29.05 &26.93 & 24.44& 23.04& -  \\ 
     & CiaoSR \cite{cao2023ciaosr} & 34.22 &30.65 &\textcolor{blue}{28.93} & 26.79 &25.28& \textcolor{red}{23.37} & 32.41 &29.34 & \textcolor{blue}{27.83} & \textcolor{blue}{26.07} &  \textcolor{blue}{25.00} & \textcolor{blue}{23.64} & 33.30& \textcolor{blue}{29.17} & \textcolor{red}{27.11} & \textcolor{blue}{24.58} & \textcolor{red}{23.13} &\textcolor{blue}{21.42}\\ 
     & SRNO \cite{wei2023super} & \textcolor{blue}{34.27}& \textcolor{blue}{30.71} &\textcolor{red}{28.97} &26.76& 25.26 &-  & \textcolor{blue}{32.43} & \textcolor{blue}{29.37} & \textcolor{blue}{27.83} &26.04&  24.99&  -  & \textcolor{blue}{33.33} & 29.14& 26.98 & 24.43 &23.02& -\\  \cmidrule{2-20}
    & \textbf{HIIF (ours)} & \textcolor{red}{34.29} & \textcolor{red}{30.76} & 28.92 & \textcolor{red}{26.84} & \textcolor{blue}{25.32} & \textcolor{blue}{23.34} & \textcolor{red}{32.51} & \textcolor{red}{29.42} & \textcolor{red}{27.86} & \textcolor{red}{26.08} & \textcolor{red}{25.02} & \textcolor{red}{23.69} & \textcolor{red}{33.34} & \textcolor{red}{29.20} & \textcolor{blue}{27.07} & \textcolor{red}{24.59} & \textcolor{blue}{23.11} & \textcolor{red}{21.43} \\ \midrule

\multirow{8}{*}{\rotatebox{270}{SwinIR\cite{liang2021swinir}}} & SwinIR only & 34.14 &30.77 &28.94  & - & - & - & 32.44 &29.37& 27.83 & - &- &- &33.40& 29.29 &27.07 & - &- &-\\ \cmidrule{2-20}
     & MetaSR \cite{hu2019meta} & 34.14 &30.66 &28.85  & 26.58 &25.09 &23.33 &32.39 &29.31 &27.75  &25.94 &24.87& 23.59 & 33.29 &29.12 &26.76 &24.16& 22.75& 21.31\\ 
     & LIIF \cite{chen2021learning} & 34.14 &30.75& 28.98  & 26.82& 25.34& 23.38   & 32.39 &29.34 &27.84  & 26.07& 25.01& 23.64 &  33.36& 29.33 &27.15& 24.59& 23.14& 21.43 \\ 
     & LTE \cite{lee2022local} & 34.25& 30.80 & \textcolor{blue}{29.06}  & 26.86 &25.42 & \textcolor{blue}{23.44}   & 32.44& 29.39 &27.86  & 26.09& 25.03& 23.66  &33.50& 29.41& 27.24 & 24.62& 23.17& 21.50 \\ 
     & CLIT \cite{chen2023cascaded} &34.27& \textcolor{blue}{30.85} & \textcolor{red}{29.08} & \textcolor{blue}{26.94} & \textcolor{blue}{25.55} & -  & 32.46& \textcolor{blue}{29.42} & \textcolor{blue}{27.91} & \textcolor{blue}{26.15} & \textcolor{red}{25.09} & - &33.56& 29.43 & 27.25 & 24.77& 23.33& - \\ 
     & CiaoSR \cite{cao2023ciaosr} & \textcolor{blue}{34.33} &30.82 &\textcolor{red}{29.08} & 26.88 &25.42 &23.38 & \textcolor{blue}{32.47} & \textcolor{blue}{29.42} & 27.90  &26.13& \textcolor{blue}{25.07} & \textcolor{blue}{23.68} & \textcolor{blue}{33.65} & \textcolor{red}{29.52} &\textcolor{red}{27.42} & \textcolor{blue}{24.84} & \textcolor{blue}{23.34} & \textcolor{red}{21.60} \\   \cmidrule{2-20}
    & \textbf{HIIF (ours)} & \textcolor{red}{34.42} & \textcolor{red}{30.86} & 29.02 & \textcolor{red}{26.95} & \textcolor{red}{25.59} & \textcolor{red}{23.45} & \textcolor{red}{32.55} & \textcolor{red}{29.46} & \textcolor{red}{27.92} & \textcolor{red}{26.16} & \textcolor{blue}{25.07} & \textcolor{red}{23.69} & \textcolor{red}{33.67} & \textcolor{blue}{29.47} & \textcolor{blue}{27.38} & \textcolor{red}{24.86} & \textcolor{red}{23.36} & \textcolor{blue}{21.54} \\  \bottomrule
% Add additional model rows here
\end{tabular}}
\label{tbl:results2}
\end{table*}

%-------------------------------------------------------------------------
\subsection{Experiment setup}

\textbf{Dataset.} Following previous work \cite{lim2017enhanced, chen2021learning}, we use the DIV2K training dataset \cite{agustsson2017ntire} from the NTIRE 2017 Challenge \cite{timofte2017ntire} for network optimization, which consists of 800 images in 2K resolution. For evaluation, we follow common practice in continuous super-resolution and employ the DIV2K validation set (containing 100 images) and four other commonly used test sets, Set5 \cite{bevilacqua2012low}, Set14 \cite{zeyde2012single}, BSD100 \cite{martin2001database}, and Urban100 \cite{huang2015single}.

\noindent\textbf{Traning material.} According to \cite{lim2017enhanced}, we generate 48 $\times$ 48 training 
 patches from images of DIV2K training set. For arbitrary-scale down-sampling, we adopt the method described in \cite{chen2021learning, lee2022local}, and sample $B_s$ random scaling factors $r_{1\cdots B_s}$ from a uniform distribution $\mathcal{U}(1,4)$, i.e. in-scale. Here, $B_s$ is the batch size. In order to facilitate the training, we use the same scale factor for height and width, i.e. $r_x = r_y = r$, which is employed to crop ${48r \times 48r}$ patches from original images, and generate their corresponding 48 $\times$ 48 down-sampled counterparts through bicubic resizing \cite{paszke2019pytorch}. For the ground-truth images, we converted them into pixel samples (coordinate-RGB value pairs), sampling $48^2$ pixel samples from each image to standardize the ground-truth shapes within each batch. 

\noindent\textbf{Encoder backbone.} Following the practice in the previous studies \cite{chen2021learning, lee2022local}, we integrate our methods with three encoder backbones, including two CNN-based models, EDSR\_baseline \cite{lim2017enhanced} and RDN \cite{zhang2018residual} and a transformer-based encoder, SwinIR \cite{liang2021swinir}, all of which have been modified by removing their up-sampling modules.

\noindent\textbf{Implementation details.} Based on \cite{chen2021learning, lee2022local}, EDSR\_baseline and RDN based models have been trained for 1000 epochs with a batch size of 16, an initial learning rate of 1e-4, and a decay factor of 0.5 applied every 200 epochs. SwinIR based models have been trained for 1000 epochs but with a batch size of 32, an initial learning rate of 2e-4, and a decay factor of 0.5 at epochs 500, 800, 900, and 950. We use L1 loss and the Adam \cite{kingma2014adam} optimization during training. The training and testing are implemented based on the NVIDIA RTX 4090 graphic card. The hyperparameters in the proposed HIIF decoder include the level of multi-scale grids $L=6$, the inter-channel number $C=256$, and the mod factor used for hierarchical encoding $S=2$, the number of multi-head attention blocks $B=2$ and the multi-head number $N=16$. 

%%%%%%%%%%%%%%% visual com

\begin{figure*}[!t]
  \centering

    % \rotatebox{90}{\footnotesize}
        \begin{minipage}{0.245\linewidth}
		  \centering
            \setlength{\abovecaptionskip}{0.cm}
		  \includegraphics[width=1\linewidth,height=0.65\linewidth]{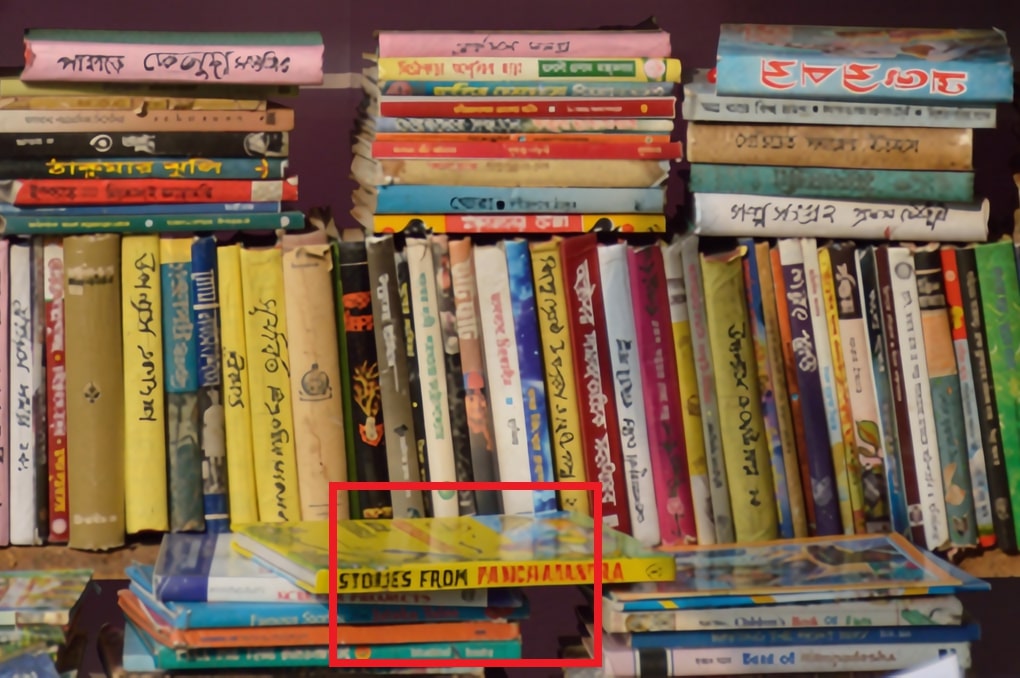}
            \caption*{\footnotesize 0867 (DIV2K$, \times$4)}
		% \caption*{GT \protect\\ (PSNR/SSIM)}
	  \end{minipage}
		\begin{minipage}{0.245\linewidth}
		  \centering
            \setlength{\abovecaptionskip}{0.cm}
		  \includegraphics[width=1\linewidth,height=0.65\linewidth]{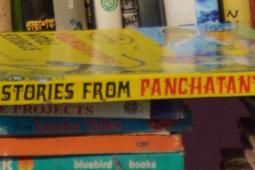}
            \caption*{\footnotesize HR }
		% \caption*{GT \protect\\ (PSNR/SSIM)}
	  \end{minipage}
        \begin{minipage}{0.245\linewidth}
		  \centering
            \setlength{\abovecaptionskip}{0.cm}
		  \includegraphics[width=1\linewidth,height=0.65\linewidth]{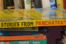}
            \caption*{\footnotesize LR }
		% \caption*{GT \protect\\ (PSNR/SSIM)}
	  \end{minipage}
	  \begin{minipage}{0.245\linewidth}
		  \centering
            \setlength{\abovecaptionskip}{0.cm}
		  \includegraphics[width=1\linewidth,height=0.65\linewidth]{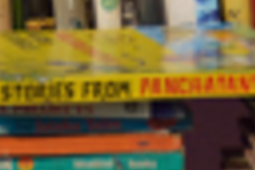}
            \caption*{\footnotesize Bicubic}
		% \caption*{EDSR\_baseline \protect\\ (35.27/0.9780)}
	  \end{minipage}
   
 	\begin{minipage}{0.245\linewidth}
		  \centering
            \setlength{\abovecaptionskip}{0.cm}
		  \includegraphics[width=1\linewidth,height=0.65\linewidth]{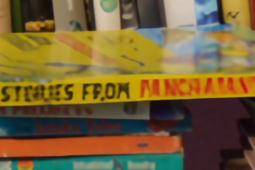}
            \caption*{\footnotesize LIIF}
		% \caption*{EDSRmini \protect\\ (34.33/0.9736)}
	  \end{minipage}
 	\begin{minipage}{0.245\linewidth}
		  \centering
            \setlength{\abovecaptionskip}{0.cm}
		  \includegraphics[width=1\linewidth,height=0.65\linewidth]{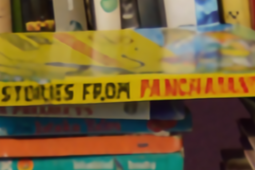}
            \caption*{\footnotesize LTE}
		% \caption*{SwinIR\_lightweight \protect\\ (35.61/0.9795)}
	  \end{minipage}
        \begin{minipage}{0.245\linewidth}
		  \centering
            \setlength{\abovecaptionskip}{0.cm}
	      \includegraphics[width=1\linewidth,height=0.65\linewidth]{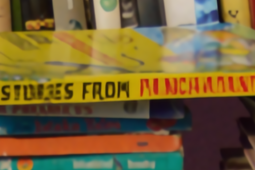}
            \caption*{\footnotesize SRNO}
		% \caption*{SwinIRmini \protect\\ (34.84/0.9901)}
	  \end{minipage}
  	\begin{minipage}{0.245\linewidth}
		  \centering
            \setlength{\abovecaptionskip}{0.cm}
	      \includegraphics[width=1\linewidth,height=0.65\linewidth]{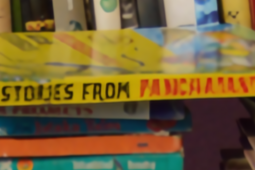}
            \caption*{\footnotesize ours}
		% \caption*{SwinIRmini \protect\\ (34.84/0.9901)}
	  \end{minipage}

   % \rotatebox{90}{\footnotesize }
        \begin{minipage}{0.245\linewidth}
		  \centering
            \setlength{\abovecaptionskip}{0.cm}
		  \includegraphics[width=1\linewidth,height=0.65\linewidth]{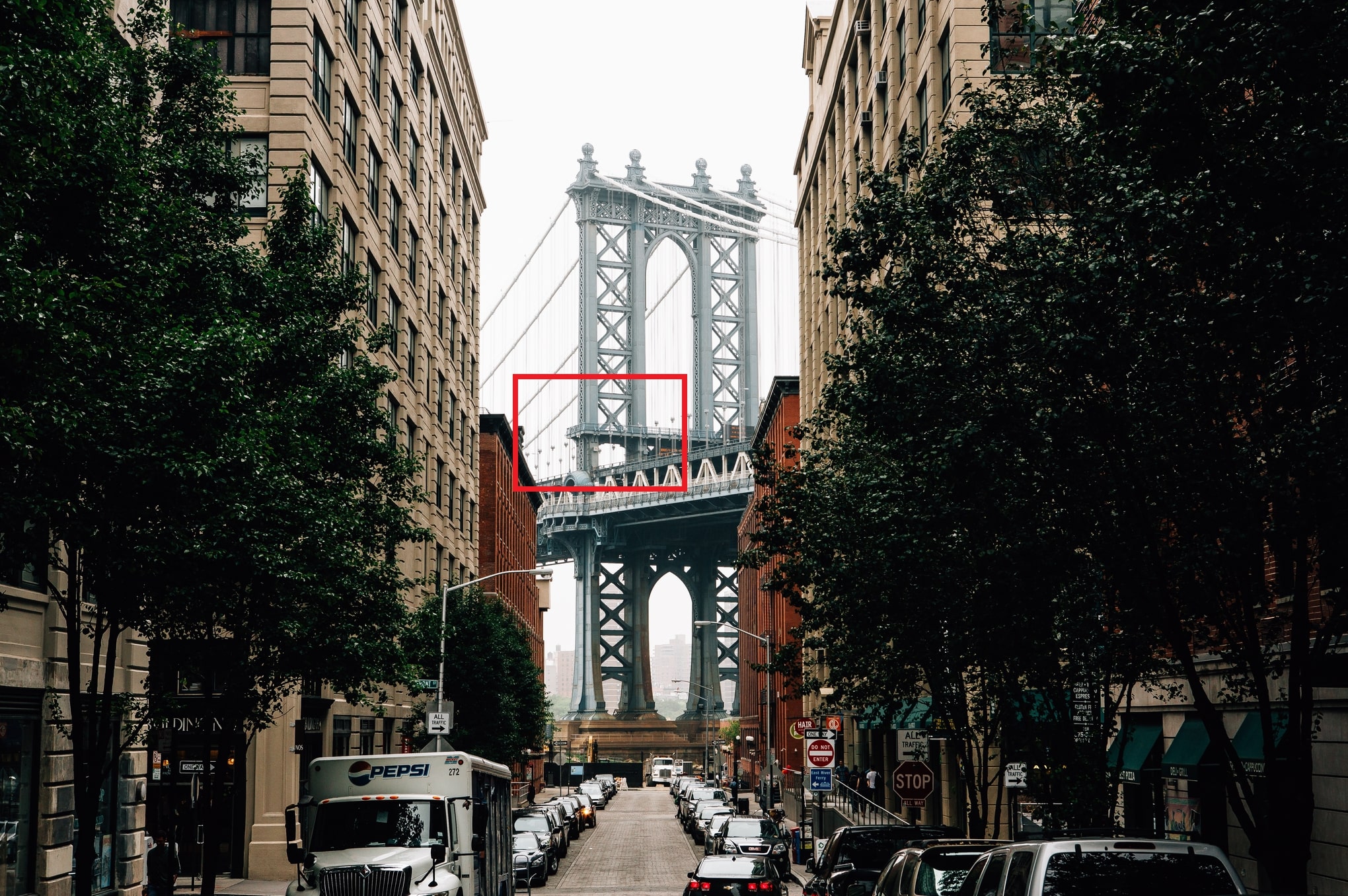}
            \caption*{\footnotesize 0861 (DIV2K, $\times$6)}
		% \caption*{GT \protect\\ (PSNR/SSIM)}
	  \end{minipage}
		\begin{minipage}{0.245\linewidth}
		  \centering
            \setlength{\abovecaptionskip}{0.cm}
		  \includegraphics[width=1\linewidth,height=0.65\linewidth]{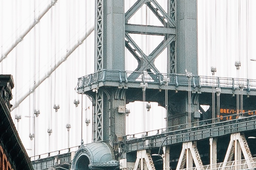}
            \caption*{\footnotesize HR }
		% \caption*{GT \protect\\ (PSNR/SSIM)}
	  \end{minipage}
        \begin{minipage}{0.245\linewidth}
		  \centering
            \setlength{\abovecaptionskip}{0.cm}
		  \includegraphics[width=1\linewidth,height=0.65\linewidth]{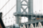}
            \caption*{\footnotesize LR }
		% \caption*{GT \protect\\ (PSNR/SSIM)}
	  \end{minipage}
	  \begin{minipage}{0.245\linewidth}
		  \centering
            \setlength{\abovecaptionskip}{0.cm}
		  \includegraphics[width=1\linewidth,height=0.65\linewidth]{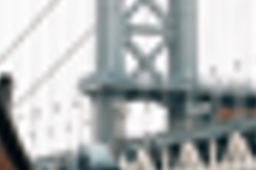}
            \caption*{\footnotesize Bicubic}
		% \caption*{EDSR\_baseline \protect\\ (35.27/0.9780)}
	  \end{minipage}
   
 	\begin{minipage}{0.245\linewidth}
		  \centering
            \setlength{\abovecaptionskip}{0.cm}
		  \includegraphics[width=1\linewidth,height=0.65\linewidth]{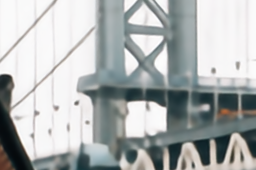}
            \caption*{\footnotesize LIIF}
		% \caption*{EDSRmini \protect\\ (34.33/0.9736)}
	  \end{minipage}
 	\begin{minipage}{0.245\linewidth}
		  \centering
            \setlength{\abovecaptionskip}{0.cm}
		  \includegraphics[width=1\linewidth,height=0.65\linewidth]{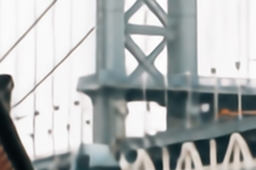}
            \caption*{\footnotesize LTE}
		% \caption*{SwinIR\_lightweight \protect\\ (35.61/0.9795)}
	  \end{minipage}
        \begin{minipage}{0.245\linewidth}
		  \centering
            \setlength{\abovecaptionskip}{0.cm}
	      \includegraphics[width=1\linewidth,height=0.65\linewidth]{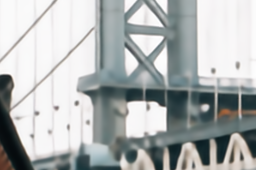}
            \caption*{\footnotesize SRNO}
		% \caption*{SwinIRmini \protect\\ (34.84/0.9901)}
	  \end{minipage}
  	\begin{minipage}{0.245\linewidth}
		  \centering
            \setlength{\abovecaptionskip}{0.cm}
	      \includegraphics[width=1\linewidth,height=0.65\linewidth]{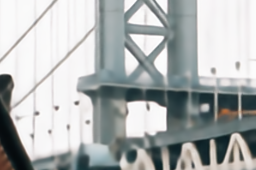}
            \caption*{\footnotesize ours}
		% \caption*{SwinIRmini \protect\\ (34.84/0.9901)}
	  \end{minipage}

  \caption{Qualitative comparison results. Here RDN \cite{zhang2018residual} is used as an encoder for all methods.}
  \label{fig:qualitative1}
\end{figure*}

%-------------------------------------------------------------------------
\subsection{Benchmark results}

To quantitatively assess the effectiveness of the proposed method, we evaluate both up-sampling tasks within the training scale distribution (i.e. $\times$2, $\times$3, and $\times$4) and those with larger scales outside this distribution (i.e. from $\times$6 to $\times$30). Specifically, for scales $\times$2, $\times$3, and $\times$4, we use the low-resolution inputs provided in the test datasets. For scales ranging from $\times$6 to $\times$30, we first crop the ground-truth images to ensure that their dimensions are divisible by the scaling factor, and then generate low-resolution inputs via bicubic down-sampling, following \cite{chen2021learning, lee2022local}. Some results are sourced from original research papers or open-source implementations.

\begin{figure}[!t]
  \centering

    % \rotatebox{90}{\footnotesize Urban100}
        \begin{minipage}{0.235\linewidth}
		  \centering
            \setlength{\abovecaptionskip}{0.cm}
		  \includegraphics[width=1\linewidth]{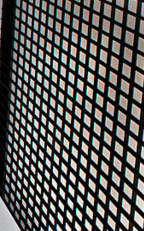}
            \caption*{\footnotesize img\_019 GT}
		% \caption*{GT \protect\\ (PSNR/SSIM)}
	  \end{minipage}
		\begin{minipage}{0.235\linewidth}
		  \centering
            \setlength{\abovecaptionskip}{0.cm}
		  \includegraphics[width=1\linewidth]{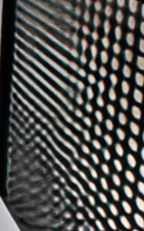}
            \caption*{\footnotesize EDSR\_bl \cite{lim2017enhanced} }
		% \caption*{GT \protect\\ (PSNR/SSIM)}
	  \end{minipage}
        \begin{minipage}{0.235\linewidth}
		  \centering
            \setlength{\abovecaptionskip}{0.cm}
		  \includegraphics[width=1\linewidth]{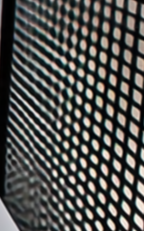}
            \caption*{\footnotesize RDN \cite{zhang2018residual} }
		% \caption*{GT \protect\\ (PSNR/SSIM)}
	  \end{minipage}
	  \begin{minipage}{0.235\linewidth}
		  \centering
            \setlength{\abovecaptionskip}{0.cm}
		  \includegraphics[width=1\linewidth]{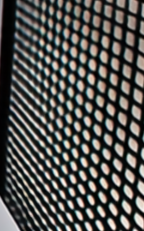}
            \caption*{\footnotesize SwinIR \cite{liang2021swinir}}
		% \caption*{EDSR\_baseline \protect\\ (35.27/0.9780)}
	  \end{minipage}

  \caption{Qualitative comparison among three encoders with HIIF for $\times 6$ upsampling. The sequence is from the Urban100 dataset.}
  \label{fig:encodercomp}
  \vspace{-5pt}
\end{figure}

\begin{figure*}[!t]
  \centering

    % \rotatebox{90}{\footnotesize }
        \begin{minipage}{0.245\linewidth}
		  \centering
            \setlength{\abovecaptionskip}{0.cm}
		  \includegraphics[width=1\linewidth]{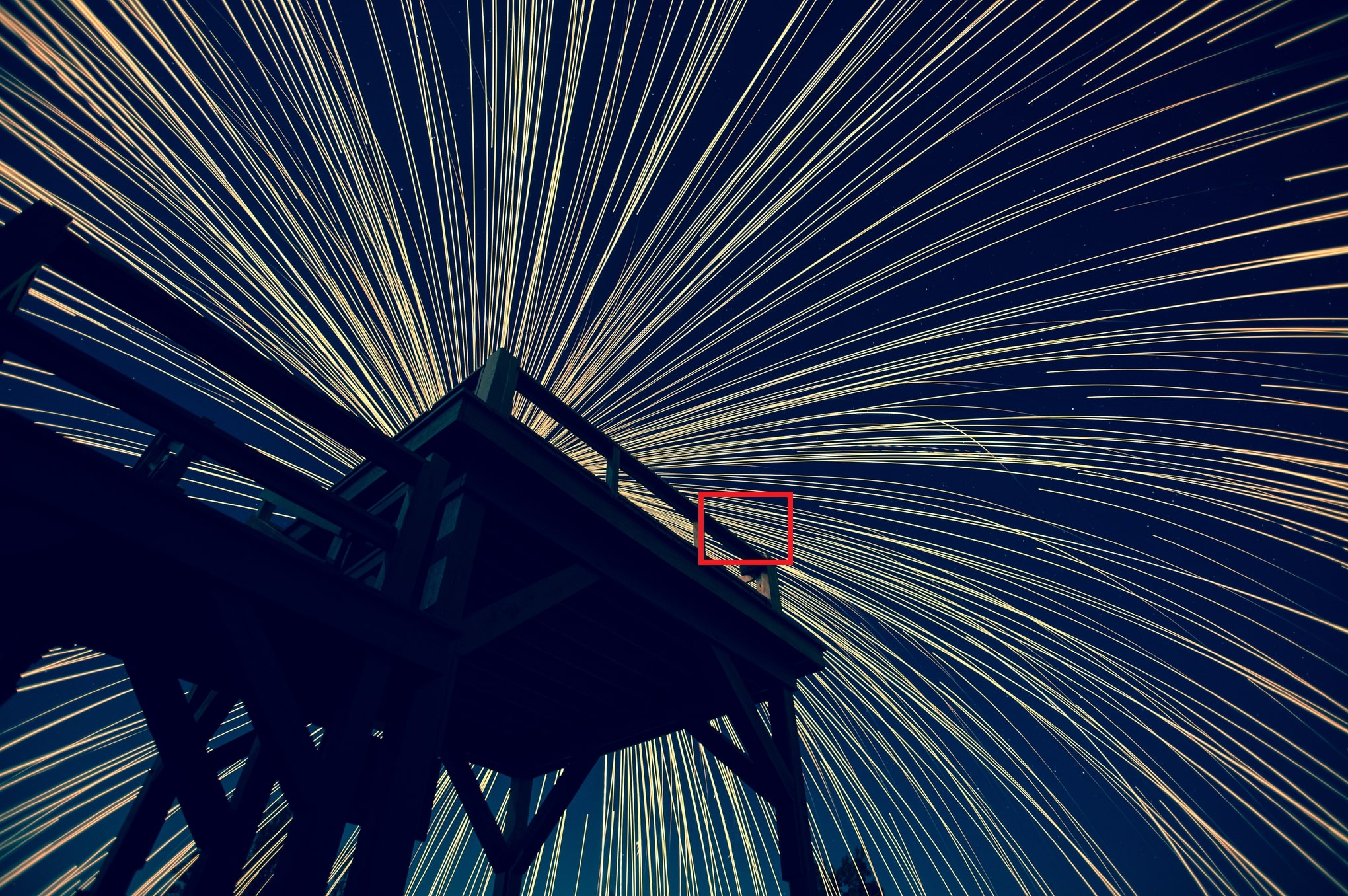}
            \caption*{\footnotesize 0828 (DIV2K, $\times$3.3)}
		% \caption*{GT \protect\\ (PSNR/SSIM)}
	  \end{minipage}
		\begin{minipage}{0.245\linewidth}
		  \centering
            \setlength{\abovecaptionskip}{0.cm}
		  \includegraphics[width=1\linewidth]{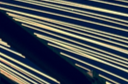}
            \caption*{\footnotesize HR }
		% \caption*{GT \protect\\ (PSNR/SSIM)}
	  \end{minipage}
        \begin{minipage}{0.245\linewidth}
		  \centering
            \setlength{\abovecaptionskip}{0.cm}
		  \includegraphics[width=1\linewidth]{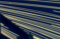}
            \caption*{\footnotesize LR }
		% \caption*{GT \protect\\ (PSNR/SSIM)}
	  \end{minipage}
	  \begin{minipage}{0.245\linewidth}
		  \centering
            \setlength{\abovecaptionskip}{0.cm}
		  \includegraphics[width=1\linewidth]{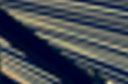}
            \caption*{\footnotesize Bicubic}
		% \caption*{EDSR\_baseline \protect\\ (35.27/0.9780)}
	  \end{minipage}
   
 	\begin{minipage}{0.245\linewidth}
		  \centering
            \setlength{\abovecaptionskip}{0.cm}
		  \includegraphics[width=1\linewidth]{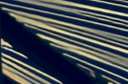}
            \caption*{\footnotesize LIIF}
		% \caption*{EDSRmini \protect\\ (34.33/0.9736)}
	  \end{minipage}
 	\begin{minipage}{0.245\linewidth}
		  \centering
            \setlength{\abovecaptionskip}{0.cm}
		  \includegraphics[width=1\linewidth]{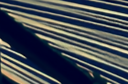}
            \caption*{\footnotesize LTE}
		% \caption*{SwinIR\_lightweight \protect\\ (35.61/0.9795)}
	  \end{minipage}
        \begin{minipage}{0.245\linewidth}
		  \centering
            \setlength{\abovecaptionskip}{0.cm}
	      \includegraphics[width=1\linewidth]{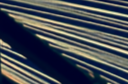}
            \caption*{\footnotesize SRNO}
		% \caption*{SwinIRmini \protect\\ (34.84/0.9901)}
	  \end{minipage}
  	\begin{minipage}{0.245\linewidth}
		  \centering
            \setlength{\abovecaptionskip}{0.cm}
	      \includegraphics[width=1\linewidth]{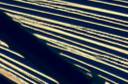}
            \caption*{\footnotesize ours}
		% \caption*{SwinIRmini \protect\\ (34.84/0.9901)}
	  \end{minipage}

  \caption{Qualitative comparison on non-integer scales. Here EDSR\_baseline \cite{lim2017enhanced} is used as an encoder for all methods.}
  \label{fig:qualitative2}
\end{figure*}

\noindent\textbf{Quantitative results.} Tables \ref{tbl:results1} and \ref{tbl:results2} summarize the quantitative results, comparing the proposed HIIF to existing arbitrary-scale SR methods, including MetaSR \cite{hu2019meta}, LIIF \cite{chen2021learning}, LTE \cite{lee2022local}, CLIT \cite{chen2023cascaded}, CiaoSR \cite{cao2023ciaosr} and SRNO \cite{wei2023super}. We test three encoder backbones on five test datasets in terms of various upsampling factors ranging from $\times 2$ to $\times 30$. It can be observed that our HIIF model consistently achieves excellent super-resolution performance (up to 0.17dB compared to the second best performer) for all scale factors, encoder backbones on five datasets among all tested continuous super-resolution methods. The results have also been illustrated by the radar plots in Figure \ref{fig:curves}. Moreover, we also provide quantitative results, directly comparing the performance between original encoder models and their integrated HIIF versions for three in-distribution scales (Figure \ref{fig:curves}). Here encoder only models were trained separately, while HIIF methods only optimize a single network to deal with super-resolution tasks for any up-sampling scales.

\noindent\textbf{Qualitative results.} Visual comparisons between HIIF and other continuous SR methods are provided in Figure \ref{fig:qualitative1} and \ref{fig:qualitative2} - the former presents the sample frames for integer scales while the latter shows those for non-integer scales. Here we compare the output of the Bicubic filter, HIIF and three SoTA models, LIIF \cite{chen2021learning}, LTE \cite{lee2022local} and SRNO \cite{wei2023super}. The results are based on EDST\_baseline and RDN encoders. It can be observed that our HIIF model offers better reconstruction results compared to the benchmark methods, with fewer blocky or structural artifacts.

\begin{table}[!t]
\centering
\caption{Comparisons of model size (M), inference time (s), and inference GPU memory (GB). Here the average time and GPU memory are measured based on an NVIDIA RTX 4090 24GB Graphic card. Here Urban100 is used as the test set.}
\resizebox{\linewidth}{!}{\begin{tabular}{r|c|c|c}
\toprule 
Encoder/Method & \#Params (M)&Runtime (s) & Memory (GB)\\ \midrule
  EDSR\_baseline \cite{lim2017enhanced} & 1.5 & 3.23 & 2.2 \\ \midrule  

  + MetaSR \cite{hu2019meta} & + 0.45 & 8.23 & 1.2  \\ 
       + LIIF \cite{chen2021learning} & +  0.35 & 18.48 & 1.3 \\ 
       + LTE \cite{lee2022local} & +  0.49 & 18.54 & 1.4 \\ 
       + CLIT \cite{chen2023cascaded} & +  15.7 & 398.02 & 16.3 \\ 
       + CiaoSR \cite{cao2023ciaosr} & +  1.43 & 251.80 & 12.6 \\ 
       + SRNO \cite{wei2023super} & +  0.81 & 20.23 & 7.1 \\ 
      \midrule
       + \textbf{HIIF (Ours)} & + 1.33 & 35.17 & 1.5  \\  \midrule
         RDN \cite{zhang2018residual} & 21.9 & 8.98 &  3.0\\ 
     SwinIR \cite{liang2021swinir} & 11.8 & 68.15 & 3.0 \\ \bottomrule
\end{tabular}}
\label{tbl:params}
\vspace{-5pt}
\end{table}

In addition, a qualitative comparison for $\times 6$ ISR task is shown in Figure \ref{fig:encodercomp}, when HIIF is integrated with three different encoders. Due to the robust reconstruction capability of the SwinIR model, it offers better image reconstruction results compared to those based on EDSR and RDN.

\noindent\textbf{Complexity analysis.} To fully investigate the characteristics of the proposed method, we report and compare its complexity in Table \ref{tbl:params} in terms of model size, inference runtime and memory usage for different continuous SR methods based on the EDSR\_baseline encoder. It can be observed that, when integrated with EDSR, HIIF results in slightly increased total model size, slower runtime and larger memory usage compared to MetaSR, LIIF, and LTE, while the complexity figures for CLIT and CiaoSR are even higher.

%-------------------------------------------------------------------------
\subsection{Ablation Study}
We conducted ablation studies to investigate the main contributions in our HIIF framework: hierarchical encoding, multi-scale architecture, and multi-head linear attention. Specifically, we created the following model variants: (v1-H) HIIF without the hierarchical encoding; (v2-MS) without the multi-scale architecture, i.e., inputting all hierarchical encodings at the beginning; (v3-MH) HIIF without the multi-head linear attention blocks. All these models are based on the EDSR-baseline encoder \cite{lim2017enhanced} and the Urban100 dataset. The ablation study results, as presented in Table \ref{tbl:ablationstudy}, show that the performance of all three variants is worse than that of the original HIIF, indicating the effectiveness of three major contributions.

\begin{table}[!t]
\centering
\caption{Ablation study results on the Urban100 dataset. Here EDSR\_baseline \cite{lim2017enhanced} is employed as the encoder.}
\resizebox{\linewidth}{!}{\begin{tabular}{r|ccc|ccc}
\toprule
% \multirow{3}{*}{\rotatebox{270}{Encoder}}&\multicolumn{1}{c|}{Database}& \multicolumn{6}{c|}{Set14} & \multicolumn{6}{c|}{BSD100} &\multicolumn{6}{c}{Urban100}\\
% \cmidrule{2-20}
\multicolumn{1}{c|}{PSNR (dB)$\uparrow$} & \multicolumn{3}{c|}{In-distribution} & \multicolumn{3}{c}{Out-of-distribution}  \\ \midrule
  Method & $\times 2$ & $\times 3$ & $\times 4$ & $\times 6$ & $\times 8$ & $\times 12$\\ \midrule
 EDSR only & 31.99 & 28.15 & 26.03 & - & - & -  \\ 
 \midrule
 v1-H & 32.56 & 28.47 & 26.38  & 23.91 & 22.59 & 21.00 \\ 
 v2-MS & 32.47 & 28.45 & 26.35  & 23.92 & 22.56 & 20.98 \\ 
 v3-MH & 32.34 & 28.34 & 26.25  & 23.86 & 22.52 & 20.96 \\ \midrule
 \textbf{HIIF (ours)} & \textbf{32.69} & \textbf{28.59} & \textbf{26.51} & \textbf{23.99} & \textbf{22.72} & \textbf{21.09} \\ \bottomrule
% Add additional model rows here
\end{tabular}}
\label{tbl:ablationstudy}
\vspace{-5pt}
\end{table}

% %-------------------------------------------------------------------------
% \subsection{Discussion}

%% file: sec/5_Conclusion.tex
\section{Conclusion}
\label{sec:Conclusion}
In this paper, we propose a novel \textbf{H}ierarchical encoding based \textbf{I}mplicit \textbf{I}mage \textbf{F}unction for continuous image super-resolution, \textbf{HIIF}. It encodes relative positional information hierarchically and local features at multiple scales, which improves connectivity between sampling points in local regions, thus strengthening the representation capabilities. Additionally, the framework integrates a multi-head self-attention mechanism within the representation network, expanding the receptive field to effectively capture non-local information. Based on the experimental results, both quantitative (up to 0.17dB in PSNR) and qualitative comparisons demonstrate the superior performance of HIIF over existing methods. Our ablation study also validates the effectiveness of the new design features introduced in this work. 